\documentclass[letterpaper, 10 pt, conference]{ieeeconf}  

\IEEEoverridecommandlockouts                              

\usepackage{amsmath} 
\usepackage{amssymb}  
\usepackage{adjustbox}
\usepackage{gensymb}
\usepackage{threeparttable}
\usepackage{multirow}
\usepackage{booktabs}
\usepackage[table]{xcolor}
\usepackage[scr=boondox,cal=esstix]{mathalpha}
\usepackage{tikz}
\usetikzlibrary{fadings}
\usepackage{pifont}
\usepackage{makecell}
\usepackage{url}
\usepackage{transparent}
\usepackage{colortbl}

\newcommand{\gradienttext}[3]{%
  \begin{tikzfadingfrompicture}[name=temp fading]
    \node[text=transparent!0, inner sep=0pt, outer sep=0pt] {#3};
  \end{tikzfadingfrompicture}%
  \tikz[baseline=(textnode.base)]{
    \node[inner sep=0pt, outer sep=0pt] (textnode) {\phantom{#3}};
    \shade[path fading=temp fading, fit fading=false,
           left color=#1, right color=#2] 
      ([xshift=-0.5pt,yshift=-0.5pt]textnode.south west) 
      rectangle ([xshift=0.5pt,yshift=0.5pt]textnode.north east);
  }%
}

\title{\LARGE \bf
\gradienttext{red}{blue}{nuTruck}: Benchmarking Autonomous Driving Planning for \\
Distributed Electric-drive Trucks 
}

\author{Jinyu Miao${}^{1}$, Pu Zhang${}^{2}$, Yifei He${}^{1}$, Chengyao Zhang${}^{1}$, Kun Jiang${}^{1,*}$, Ke Wang${}^{2}$, Mengmeng Yang${}^{1,*}$, \\Diange Yang${}^{1,*}$ 
\thanks{This work was supported in part by the National Natural Science Foundation of China (52394264, 52472449, U22A20104, 52372414, 52402499), and China Postdoctoral Science Foundation (2024M761636).}
\thanks{$^{1}$Jinyu Miao, Yifei He, Chengyao Zhang, Kun Jiang, Mengmeng Yang, and Diange Yang are with the School of Vehicle and Mobility, Tsinghua University, Beijing, China. {\tt\small jinyu.miao97@gmail.com}}%
\thanks{$^{2}$Pu Zhang, and Ke Wang are with KargoBot.AI Inc., Beijing, China}%
\thanks{$^{*}$Corresponding author: Diange Yang, Kun Jiang, and Mengmeng Yang}%
}

\begin{document}

\maketitle
\thispagestyle{empty}
\pagestyle{empty}

\begin{abstract}
The dominance of traditional rule-based methods in autonomous driving has gradually been replaced by learning-based approaches. While learning-based planners have achieved considerable success in passenger vehicles, their performance on heavy-duty trucks, particularly modern distributed electric-drive trucks (DETs), remains largely unexplored. To facilitate research and application of learning-based planners in DETs, this letter presents the first high-fidelity benchmark, called nuTruck, designed to support large-scale neural network training and closed-loop evaluation. Given the complex dynamics and high rollover susceptibility of DETs, we first incorporate a highly accurate nonlinear truck dynamical model into the simulation, which enables independent driving and steering of all wheels and captures dynamic load transfer caused by acceleration, deceleration, and cornering, thereby allowing quantitative assessment of rollover risk in closed-loop simulation. Second, we adapt several rule-based and learning-based planners as baselines for DETs and evaluate their performance in closed-loop simulation. Finally, using real-world driving scenarios from the nuPlan dataset, we conduct extensive closed-loop evaluations, analyzing not only conventional collision-free planning performance, but also the dynamical safety of the planned trajectories. The proposed nuTruck benchmark is expected to serve as a new standard for fair and realistic evaluation of autonomous driving planners on DETs.

\end{abstract}

\section{Introduction}
\label{sec:intro}

Autonomous driving (AD) technology has garnered worldwide attention and achieved remarkable success over recent decades. The effectiveness of modern AD systems largely relies on massive amounts of real-world data, which has accelerated the transition from traditional rule-based methods \cite{apollo,RoboCar} towards learning-based approaches \cite{hu2023_uniad,Sun2024SparseDriveEA}. Rather than manually designing heuristic rules and tuning hyperparameters through trial-and-error approaches, modern AD algorithms learn from data, offering promising potential for improving planning modules that provide reference trajectories for vehicles.

Learning-based planners demand high-fidelity simulation that reflects the complexity of real-world driving. Datasets such as nuScenes \cite{nuscenes} and Waymo \cite{Ettinger_2021_ICCV} provide human driving data to support imitation learning (IL)-based planners \cite{hu2023_uniad,Sun2024SparseDriveEA}, but they only support open-loop evaluation that ignores interactions with surrounding agents \cite{nuplan}, tracking error accumulation \cite{Cheng2023RethinkingIP}, and multi-modal solutions.
To address this, closed-loop benchmarks have been proposed \cite{carla,nuplan,waymax,bench2drive}. CARLA \cite{carla} enables training and evaluation of reinforcement learning (RL)-based planners \cite{roach,li2024think} but relies on manually engineered scenarios that cannot fully replicate the complexity of real-world traffic. Bench2Drive \cite{bench2drive} improves the design of scenarios, yet still suffers from a sim-to-real gap. In contrast, nuPlan \cite{nuplan} and Waymax \cite{waymax} use real human driving logs, better reflecting real-world traffic while supporting both open-loop and closed-loop evaluation.

Despite these well-known advances for passenger cars, there are very few AD benchmarks for heavy-duty trucks. Although TruckScenes \cite{truckscenes2024} and TruckDrive \cite{Ghilotti2026TruckDriveLA} provide large-scale annotated data collected by trucks, they remain limited to open-loop evaluation and ignore the dynamical attributes of trucks, which significantly distinguishes trucks from passenger cars. For modern distributed electric-drive trucks (DETs), there is no benchmark available to support large-scale training and evaluation of learning-based planners, which hinders relevant research progress.

\begin{figure}[!t]
    \centering
    \includegraphics[width=\linewidth]{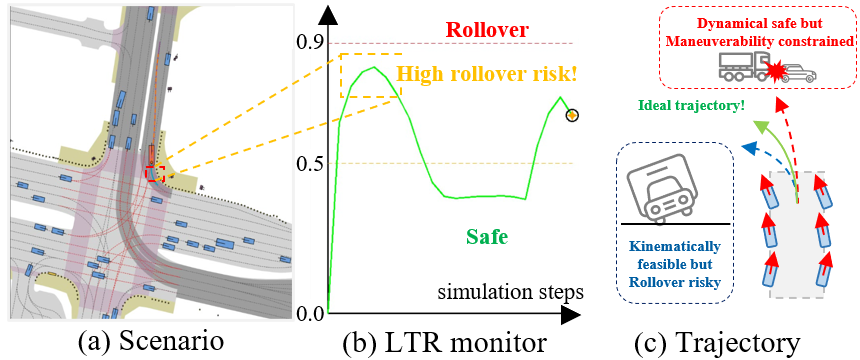}
    \caption{A closed-loop trajectory tracking case. Although the controller precisely follows a collision-free trajectory (a), the DET has high rollover risk due to an excessively small turning radius and high vehicle speed (b). An ideal trajectory for DETs should take into account the kinematical feasibility and dynamical safety (c).}
    \label{fig:intro}
\end{figure}

\begin{table}[!t]
	\centering
	\caption{The closed-loop performance of tracking collision-free trajectories in nuPlan \cite{nuplan} and nuTruck benchmarks. ''+S'' means rollover prevention and ``NRS'' is non-rollover score.}
    \renewcommand\arraystretch{1.5}
	\begin{threeparttable}
		\begin{tabular}{ l l l | c c c}
			\toprule
			Benchmark & Vehicle & Controller & CLS & CLS-Safe & NRS \\
			\midrule
            nuPlan \cite{nuplan} & Car & iLQR & 96.24 & - & - \\
            nuTruck & DET & iLQR & 77.55 & 65.75 & 83.76 \\
            nuTruck & DET & iLQR+S & 74.36 & 70.25 & 92.10 \\
			\bottomrule
		\end{tabular}
	\end{threeparttable}
	\label{table:intro}
\end{table}

\begin{table*}[!t]
	\centering
	\caption{Comparison with relevant datasets and simulations for the AD planning task.}
    \renewcommand\arraystretch{1.5}
	\begin{threeparttable}
    \resizebox{\linewidth}{!}{
		\begin{tabular}{ l l | r r r r r r r}
			\toprule
			{Dataset/Simulator} & Year & Sensor & \makecell[r]{Closed-loop\\simulation} & Driving scenarios & Agent behavior & Vehicle type & DET & Vehicle model \\
			\midrule
            nuScenes \cite{nuscenes} & 2020 & \checkmark & \ding{53} & Real-world \& Complex & Real data & Car & \ding{53} & N.A. \\
            TruckScenes \cite{truckscenes2024} & 2024 & \checkmark & \ding{53} & Real-world \& Complex & Real data & Truck & \ding{53} & N.A. \\
            TruckDrive \cite{Ghilotti2026TruckDriveLA} & 2026 & \checkmark & \ding{53} & Real-world \& Complex & Real data & Truck & \ding{53} & N.A. \\
            \hline
            CARLA \cite{carla} & 2017 & \checkmark & \checkmark & Simulated \& Medium & Inherit rules & Car, Truck & \ding{53} & Dynamical model \\
            nuPlan \cite{nuplan} & 2021 & \checkmark& \checkmark & Real-world \& Complex & Real data \& IDM \cite{idm} & Car & \ding{53} & Kinematic model \\
            Waymax \cite{waymax} & 2023 & \ding{53} & \checkmark & Real-world \& Complex & Real data \& IDM \cite{idm} & Car & \ding{53} & Kinematic model \\
            Bench2Drive \cite{bench2drive} & 2024 & \checkmark & \checkmark & Simulated \& Complex & Inherit rules & Car & \ding{53} & Dynamical model \\
            TruckSim & - & \ding{53} & \checkmark & Simulated \& Easy & Custom algorithm & Truck & \checkmark & Dynamical model \\
            \hline
            nuTruck (Ours) & 2026 & \checkmark & \checkmark & Real-world \& Complex & Real data \& IDM \cite{idm} & Truck & \checkmark & Dynamical model \\
			\bottomrule
		\end{tabular}
    }
	\end{threeparttable}
	\label{table:dataset}
\end{table*}

In addition to the absence of closed‑loop benchmarks for DETs, a further challenge in deploying learning‑based planners is that current planners fail to account for dynamical safety. DETs feature independently driven and steered wheels, significantly increasing the complexity of their dynamical model. While this enhances maneuverability, DETs have a higher center of gravity (CoG) and larger dimensions than passenger cars, making them more prone to rollovers and collisions. As noted in TruckDrive \cite{Ghilotti2026TruckDriveLA}, a learning-based planner deployed on a truck struggles to match its performance on passenger cars, even in open-loop evaluation. 
Our experiments further reveal that kinematically tracking collision-free trajectories may induce rollover risk in DETs as shown in Fig. \ref{fig:intro} (a) and (b), while considering rollover prevention only in the controller degrades the tracking accuracy and results in unsatisfactory closed-loop planning performance as shown in Tab. \ref{table:intro}.
This underscores that for DETs, AD planning algorithms must consider not only kinematic feasibility (collision-free) but also dynamical safety (rollover prevention) as shown in Fig. \ref{fig:intro} (c), generating high-quality trajectories that can be accurately tracked by a low-level controller without rollovers and collisions.

Motivated by these gaps, we propose nuTruck, a benchmark designed to enable large-scale training and closed-loop evaluation of planners specifically for DETs. Driving scenarios are constructed by filtering real-world data from nuPlan dataset \cite{nuplan}, ensuring complexity, variability, and kinematic feasibility for DETs. A high-fidelity nonlinear dynamical model of DETs is integrated into the simulation to simulate ego motion while monitoring dynamical states and rollover risk, thereby supporting quantitative assessment of dynamical safety. Furthermore, we reimplement several well-known rule-based and learning-based planners as baselines, and evaluate their performance on DETs.

In summary, the main contributions of this letter include:
\begin{itemize}
    \item We propose nuTruck, the first benchmark specifically for DETs, which supports large-scale training and testing planners in diverse real-world driving scenarios.
    \item We design a nonlinear dynamical model for DETs within the simulation to capture vehicle motion and dynamical attributes, which enables an additional evaluation metric that accounts for rollover risk, thus reflecting the dynamical safety of planned trajectories.
    \item We evaluate multiple planners in nuTruck benchmark, demonstrating their performance on DETs and highlighting the necessity of incorporating dynamical safety into the planning module. We also analyze the role of action pattern and scalability in AD planner training tailored to DETs, the derived findings are expected to motivate subsequent studies.
\end{itemize}

\begin{figure*}[!t]
    \centering
    \includegraphics[width=\linewidth]{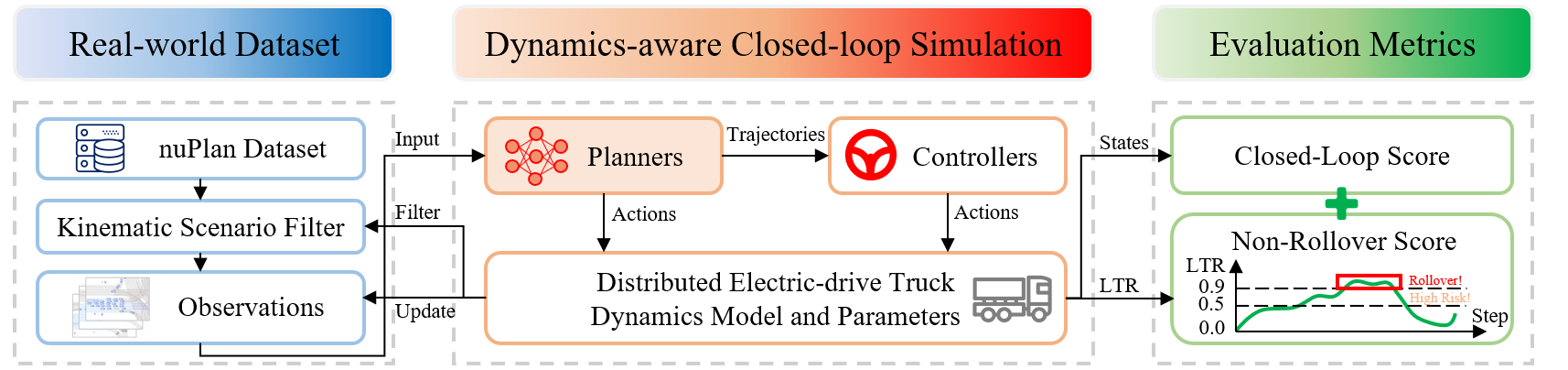}
    \caption{The overview framework of the nuTruck benchmark.}
    \label{fig:overview}
\end{figure*}

\section{Related Works}
\label{review}

In this section, we first review the relevant datasets and simulators for the AD planning task, and then introduce some well-known planning algorithms.

\subsection{Datasets and Simulators for Planning}

The success of data-driven AD algorithms relies on large-scale real-world data. 
For planning task, nuScenes \cite{nuscenes} has been widely used due to its rich expert driving logs \cite{Sun2024SparseDriveEA,hu2023_uniad}. However, it only supports open-loop evaluation, which cannot capture multi-modal solutions in complex scenarios nor account for the accumulated drift due to deviations of expert trajectories \cite{Chen2023EndtoEndAD}. TruckScenes \cite{truckscenes2024} and TruckDrive \cite{Ghilotti2026TruckDriveLA} focus on perception task for trucks, but still only enable open-loop planning evaluation.

To enable closed-loop evaluation, CARLA \cite{carla} is proposed and has supported various AD reseaches. It supports multiple vehicle types and provides built-in dynamical models. However, its manually designed scenarios and rendered appearance lead to unavoidable sim-to-real gaps. Bench2Drive \cite{bench2drive} selects more challenging scenarios based on CARLA \cite{carla} and provides expert trajectories generated by an RL-based planner \cite{li2024think}. TruckSim is a mature software with high-precision dynamical models, supporting even modern DETs. But they suffer from the same sim-to-real gap.

To overcome this, nuPlan \cite{nuplan} and Waymax \cite{waymax} were developed. They provide large-scale annotated data from human driving logs and support both open-loop and closed-loop simulation, where planned trajectories are tracked by a controller and the ego state is updated iteratively. Their scenarios are collected from real-world data, ensuring realism, complexity, and diversity. However, these two datasets do not consider dynamical models, and the vehicle motion is calculated by a kinematic bicycle model.

As summarized in Tab.~\ref{table:dataset}, existing datasets and simulators focus on traditional passenger cars and only consider kinematic feasibility, which is reasonable given the inherent stability of small vehicles. In contrast, DETs offer enhanced control flexibility but exhibit complex dynamics and a high rollover risk. This disparity motivates closed-loop simulations that integrate high-fidelity dynamical models, real-world scenarios, and realistic agent interactions, which are urgently needed for planning research on DETs.

\subsection{Learning-based Planning Algorithms}

The development paradigm of planning algorithms has shifted from hand-crafted rules to learning from massive data.
IL-based methods mimic expert driving behaviors from real-world data \cite{hu2023_uniad,Sun2024SparseDriveEA}. While scalable \cite{Naumann2025DataSL}, IL methods suffer from distribution shift between open-loop training and closed-loop inference \cite{carl}, and collecting high-quality driving data is difficult and costly.
RL-based approaches offer an alternative by optimizing planners via rewards in simulation rather than regression errors \cite{Kendall2019icra,urbandriver,li2024think}. 
A key challenge in RL methods is to design an appropriate reward function \cite{KNOX2023103829}. 
CaRL \cite{carl} trains a planner using proximal policy optimization (PPO) strategy and shows scalable improvement with more data and training steps.

Despite the success of learning-based planners for passenger cars, their application to heavy-duty trucks, especially DETs, remains limited. Wang \textit{et al.} \cite{Wang2022IITS} built a highway traffic simulation to train a RL-based planner for trucks, but they only predicted high-level driving decisions about lane changes. In TruckDrive \cite{Ghilotti2026TruckDriveLA}, the authors showed that UniAD \cite{hu2023_uniad} performs worse on a truck than on a passenger car in open-loop evaluation, suggesting that planning for trucks may be inherently more challenging.

Critically, existing AD planning and control systems typically separate responsibilities, that is, the planner ensures kinematic feasibility, while the controller handles dynamical safety. This decoupled design may fail for DETs since a kinematically feasible trajectory can be dynamically infeasible and lead to rollover as shown in Fig. \ref{fig:intro}. To address this, we propose nuTruck, a novel benchmark for training and evaluating learning-based planners on DET platforms.

\section{nuTruck benchmark}
\label{sec:method}

\subsection{Overview of the Framework}
\label{sec:overview}

To build a high-fidelity closed-loop benchmark for DETs, we propose nuTruck as shown in Fig. \ref{fig:overview}, built upon the widely recognized nuPlan benchmark \cite{nuplan} that offers high-quality driving scenarios derived from real-world human driving data. First, we integrate a nonlinear dynamical model for DETs that supports individual steering and force control for each wheel. Next, several state-of-the-art planners \cite{idm,pdm,planr1,carl} and iLQR controllers \cite{ilqr} are implemented as baselines to enable closed-loop simulation for DETs. Finally, in addition to the common closed-loop metrics, we incorporate the rollover risk of DETs as an additional evaluation metric to indicate driving safety.

\subsection{High-fidelity Dynamical Model for DET}
\label{sec:dynamics}

\subsubsection{Model defintion}
A nonlinear dynamical model of a three-axle DET featuring six independently steered and driven wheels is implemented and integrated into the closed-loop simulation.
The propagated dynamical states of the model are defined as $\textbf{s}_t=\left[v^t_x,v^t_y,\omega^t_z,p^t_x,p^t_y,\psi^t_z\right] \in \mathbb{R}^6$, where $t$ is the timestamp (index) in the simulation, $v^t_x,v^t_y$ are the longitudinal and lateral CoG velocities in the vehicle frame, $\omega^t_z$ and $\psi^t_z$ are the yaw rate and yaw angle, and $p^t_x,p^t_y$ represent the CoG position in the global frame. 
To enable distributed driving capability, the model accepts control inputs
$\textbf{u}_t={\left[T^t_{1}\sim T^t_{6},\delta^t_{1}\sim\delta^t_{6}\right]}\in \mathbb{R}^{12}$, where 
$T_i$
is the driving torque applied to the $i$-th wheel, and $\delta_i\in \left[-\pi/2,\pi/2\right]$ is the steering angle (front axle: $i=1,2$, middle axle: $i=3,4$, rear axle: $i=5,6$). 
The dynamical model can be mathematically described as a mapping function $\mathcal{F}_{\theta}:\mathbb{R}^{6} \times \mathbb{R}^{12} \rightarrow \mathbb{R}^{6}$:
\begin{equation}
    \label{equ:1}
    \dot{\textbf{s}}_t=\mathcal{F}_{\theta}(\textbf{s}_t,\textbf{u}_t)
\end{equation}
where $\dot{\textbf{s}}$ denotes the state derivatives. For notational simplicity, we omit the timestamp $t$ of the variables in the following paragraphies. Theotically, this dynamical model constitutes a typical ordinary differential equation, for which the Euler method is employed to propagate states efficiently.

\subsubsection{Tyre dynamical model}
To model the dynamical attributes of each wheel, we first compute its velocity in the vehicle frame using the vehicle states $v_x$ and $v_y$, and then obtain tyre slip angle $\alpha_i$ between the wheel velocity and the wheel plane defined by $\delta_i$. 
The cornering force of each wheel is given by a simplified Pacejka tyre magic model \cite{Pacejka01011992}:
\begin{equation}
    F_{\alpha,i}=D\text{sin}(C\text{atan}(B\hat{\alpha}_i-E(B\hat{\alpha}_i-\text{atan}(B\hat{\alpha}_i))))+S_v
\end{equation}
where $\hat{\alpha}_i=\alpha_i+S_h$ and $B,C,D,E,S_h,S_v$ denote the parameters of the tyre model

In DETs, each wheel is actuated by an individual driving torque, producing a longitudinal tyre force $F_{d,i}=T_i/R$ along the wheel plane, where $R$ is the wheel radius.
During vehicle motion, each wheel is opposed by rolling resistance, which is modeled as a linear function of the normal load $F_{z,i}$ at that wheel: $F_{r,i}=C_r\cdot F_{z,i}$, where $C_r$ denotes the rolling resistance coefficient.

Therefore, the force acting on each wheel in the vehicle frame can be summarized as:
\begin{equation}
\begin{aligned}
    F_{x,i}&=\left(F_{d,i}-F_{r,i}\right)\cdot\text{cos}(\delta_i)-F_{\alpha,i}\cdot\text{sin}(\delta_i) \\
    F_{y,i}&=\left(F_{d,i}-F_{r,i}\right)\cdot\text{sin}(\delta_i)+F_{\alpha,i}\cdot\text{cos}(\delta_i) 
\end{aligned}
\end{equation}

\subsubsection{Rigid-body dynamical model}
Incorporating all tyre forces and moments, the resulting motion in the vehicle frame is governed by:
\begin{equation}
    \begin{aligned}
        ma_x &= m(\dot{v}_x-v_y\omega_z) = \sum F_{x,i}-F_\text{drag} \\
        ma_y &= m(\dot{v}_y+v_x\omega_z) = \sum F_{y,i} \\
        I_z\dot{\omega}_z &= \sum \left(x_i F_{y,i} -y_i F_{x,i} \right)
    \end{aligned}
\end{equation}
where $m$ is the total mass of the vehicle system (including sprung, unsprung, and load components), $a_x,a_y$ are the longitudinal and lateral accelerations, $I_z$ is the total yaw moment of inertia, $x_i,y_i$ are the $i$-th wheel positions relative to CoG. Aerodynamic drag $F_\text{drag}$ is incorporated solely into the longitudinal dynamics:
\begin{equation}
    F_\text{drag}=\frac{1}{2}\rho C_d A v_x |{v_x}|
\end{equation}
where $\rho$ is air density, $C_d$ is air resistance coefficient, and $A$ is the frontal area.
The global motion of the vehicle is described by a kinematic model as follows:
\begin{equation}
\begin{aligned}
    \dot{p}_x &= v_x\text{cos}(\psi_z)-v_y\text{sin}(\psi_z) \\
    \dot{p}_y &= v_x\text{sin}(\psi_z)+v_y\text{cos}(\psi_z) \\
    \dot{\psi_z} &= \omega_z
\end{aligned}
\end{equation}

\subsubsection{Dynamic vertical loads}
To capture the dynamic influence of vehicle steering, acceleration and deceleration on vertical loads and thereby quantify rollover risk, we additionally measure the longitudinal and lateral load transfers induced by accelerations.
First, the static vertical load $F^s_{z,i}$ at each wheel is computed based on mass distribution and wheel positions. Then, given the longitudinal and lateral accelerations $a_x$ and $a_y$, the longitudinal and lateral load transfers are calculated:
\begin{equation}
    \begin{aligned}
        \sum F^{d,lon}_{z,i} &= m a_x h \\
        \sum F^{d,lat}_{z,i} &= 2m a_y h / w
    \end{aligned}
\end{equation}
where $h$ is the height of the CoG, and $w$ is the track width.
Therefore, the overall vertical load at each wheel is $F_{z,i}=F^s_{z,i}+F^{d,lon}_{z,i}+F^{d,lat}_{z,i}$, and we measure the load transfer ratio (LTR) as an indicator of rollover risk:
\begin{equation}
    \text{LTR}=\left|\frac{\sum_{i\in \text{left}} F_{z,i}-\sum_{i\in \text{right}} F_{z,i}}{\sum_{i} F_{z,i}}\right| \in [0,1]
\end{equation}
A higher LTR value indicates a greater rollover risk for the DET. In the simulation, an LTR exceeding the threshold $s_{max}$ is defined as a rollover event.

\subsection{Planning Baselines}
\label{sec:planning}

In this work, two rule-based planners \cite{idm,pdm} and five learning-based planners \cite{pluto,carl} are adopted and evaluated in the nuTruck benchmark to test its planning capability for DETs.

\subsubsection{IDM} We adopt the IDM planner \cite{idm} from the nuPlan benchmark \cite{nuplan} with hyperparameter tuning.

\subsubsection{PDM-Closed} The PDM-Closed planner \cite{pdm} evaluates sampled candidate trajectories in a closed-loop scoring stage. Accordingly, we adapt the vehicle parameters used in this stage to those of our DET, as they affect collision detection.

\subsubsection{Plan-R1} The off-the-shelf Plan-R1 planner \cite{planr1} is directly utilized due to its state-of-the-art performance on the nuPlan benchmark \cite{nuplan}. We do not fine-tune its parameters due to the lack of expert trajectory data for DETs.

\subsubsection{CaRL-A12} We train and evaluate the RL-based CaRL planner \cite{carl} within the nuTruck simulation. We expand its action prediction space to 12 dimensions, ensuring consistency with the control inputs required by DETs.

\begin{figure}[!t]
    \centering
    \includegraphics[width=\linewidth]{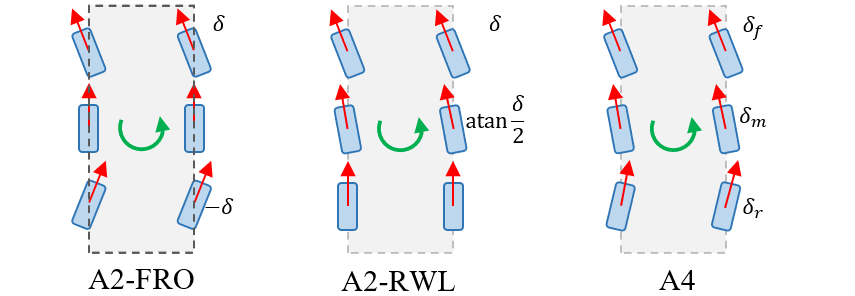}
    \caption{The action patterns used in CaRL-A2 and CaRL-A4 planners.}
    \label{fig:pattern}
\end{figure}

\subsubsection{CaRL-A2} 
Due to the inherent difficulties of distributed drive control, directly predicting high-dimensional control signals for each wheel poses challenges for learning-based planners, potentially leading to infeasible actions and impaired training convergence. To address this, we introduce two primary patterns as action priors for the CaRL planner \cite{carl}, namely, front-rear wheel opposite (FRO) and rear wheel locked (RWL) as shown in Fig. \ref{fig:pattern}. Specifically, the modified planner predicts only a single driving torque $T$ and a single steering angle $\delta$, which are subsequently expanded into 12-dimensional actions according to the following strategy:
\begin{equation}
    \begin{aligned}
        \text{FRO}&: T_{*}=T,\delta_{\text{f}}=\delta,\delta_{\text{m}}=0,\delta_{\text{r}}=-\delta \\
        \text{RWL}&: T_{*}=T,\delta_{\text{f}}=\delta,\delta_{\text{m}}=\text{atan}(\frac{\delta}{2}),\delta_{\text{r}}=0 \\
    \end{aligned}
\end{equation}
where $\delta_f,\delta_m,\delta_r$ represent the steering angles for the front, middle, and rear axles. 
In the FRO pattern, the rear wheels are steered in the opposite direction to the front wheels, thereby reducing the turning radius. 
In contrast, under the RWL pattern, the front and middle wheels follow Ackermann steering geometry, while the rear wheels are locked with no steering input. 
By employing these patterns, the action prediction space is significantly reduced, which should facilitate convergence during network training.

\subsubsection{CaRL-A4}
We also implement a variant to fully exploit the control capability of DETs, where the CaRL planner \cite{carl} predicts a four-dimensional action signal consisting of a torque $T$ and 3 steering angles $\delta_f,\delta_m,\delta_r$. The steering angles of the left and right wheels on each axle are identical, and the torque is shared equally across all wheels.

In the simulation, planners that predict trajectories \cite{idm,pdm,planr1} rely on a trajectory tracking controller to drive the truck. In contrast, the CaRL-series planners \cite{carl}, which predict actions, directly control the truck using the estimations.

\subsection{Model-based Trajectory Tracking Controller}
\label{sec:controller}

To conduct closed-loop simulation for DETs, we develop a trajectory tracking controller that estimates low-level actions (\textit{i.e.}, driving torques and steering angles for each wheel) from trajectories predicted by planners. To achieve precise trajectory tracking, the controller employs the dynamical model established in Sec. \ref{sec:dynamics} during its rollout process. Given the nonlinearity of this dynamical model, linear controllers \cite{Moser2018,Chen2019} are unsuitable. Instead, we adopt an iLQR controller \cite{ilqr} to locally linearize the dynamical model around the reference trajectories. The cost function optimized by the iLQR controller is formulated as follows:
\begin{equation}
    \mathcal{L}=\sum^{N-1}_{t=1} \mathscr{l}(\textbf{s}_t,\textbf{u}_t)+\mathscr{l}_{f}(\textbf{s}_{N})
\end{equation}
where $\mathscr{l}(\cdot,\cdot)$ is a composite function of the tracking error and the penalty on control signals:
\begin{equation}
\begin{aligned}
    \mathscr{l}(\textbf{s}_t,\textbf{u}_t)&=\frac{1}{2}{\left(\textbf{s}_t-\hat{\textbf{s}}_t\right)}^TQ\left(\textbf{s}_t-\hat{\textbf{s}}_t\right) \\
    &+\frac{1}{2}{\left(\textbf{u}_t-\textbf{u}_{0}\right)}^TR\left(\textbf{u}_t-\textbf{u}_{0}\right)
\end{aligned}
\end{equation}
where $\hat{\textbf{s}}$ is the reference trajectory from the planner, $\textbf{u}_0$ is the current control signals, and $Q$ and $R$ are the diagonal weighting matrices. $\mathscr{l}_f$ is the terminal error defined as $\mathscr{l}_f=\frac{\lambda}{2}{\left(\textbf{s}_N-\hat{\textbf{s}}_N\right)}^TQ\left(\textbf{s}_N-\hat{\textbf{s}}_N\right)$, with $\lambda>1$ serving as the terminal factor that emphasizes the final tracking errors. $N$ is the prediction horizon. The optimized control inputs are defined as four-dimensional vectors consisting of torque $T$ and steering angles $\delta_f,\delta_m,\delta_r$. These are subsequently expanded to 12-dimensional actions following the same scheme as CaRL-A4 described in Sec. \ref{sec:planning}.

In addition to this iLQR controller, we implement another variant that explicitly accounts for reducing rollover risk, denoted as iLQR-S. Specifically, a penalty term on the LTR value is incorporated into the cost function to be optimized:
\begin{equation}
\begin{aligned}
    \mathscr{l}'(\textbf{s}_t,\textbf{u}_t,\text{LTR}_t)&=\mathscr{l}(\textbf{s}_t,\textbf{u}_t)+\frac{1}{2}Q_{l}\cdot\text{LTR}_t^2\\
    &+\frac{1}{2}Q_{v}\cdot{\left(\text{LTR}_t-s_{max}\right)}^2
\end{aligned}
\end{equation}
where $Q_{l}$ and $Q_{v}$ are weighting scalars.

\subsection{Evaluation}
\label{sec:eval}

We adopt driving scenario data from the nuPlan dataset \cite{nuplan}. Since nuPlan \cite{nuplan} is collected using small-scale passenger cars, some scenarios are impassable for the large-scale DET, such as extremely crowded roads. To filter out such infeasible scenarios, we utilize the expert trajectories from nuPlan \cite{nuplan} but substitute the vehicle parameters with those of our DET. The truck then perfectly tracks these trajectories to detect scenarios in which collisions occur. Through this process, all remaining scenarios are kinematically feasible and are subsequently used for evaluation.

In the simulation, we emphasize dynamical safety for DETs in addition to the commonly used driving performance metrics from the nuPlan benchmark \cite{nuplan}. Specifically, we additionally account for rollover risk. 
The planner receives a dynamical safety metric called Non-Rollover Score (NRS) that is determined by both the maximum and mean LTR values observed during simulation. Specifically, the NRS metric is set to zero if the maximum LTR exceeds $s_{max}$. If the maximum LTR is below $s_{safe}=0.5$, the score is one. In all other cases, the score decays quadratically from one to zero as a function of the mean LTR over the scenario.

The safety score serves as a multiplicative factor when aggregated into the closed-loop metric. We refer to this modified closed-loop metric as "CLS-Safe".

\section{Experiments}

\begin{table}[!t]
	\centering
	\caption{The configuration in the nuTruck simulation.}
    \renewcommand\arraystretch{1.5}
	\begin{threeparttable}
    \resizebox{\linewidth}{!}{
		\begin{tabular}{c c | c c | c c}
			\toprule
			{Config} & Value & Config & Value & Config & Value \\
			\midrule
            $m$ & 1.051$e^4$ kg & $B$ & 6.876 & $C$ & 1.600 \\
            $D$ & 12712.300 N & $E$ & -0.2998 & $S_h$ & -0.00001 rad \\
            $S_v$ & 1.400 N & $s_{max}$ & 0.900 & $R$ & 0.510 m \\
            $C_r$ & 0.004 & $I_z$ & 6.609$e^4$ kg$\cdot m^2$ & $\rho$ & 1.206 kg/$m^3$ \\
            $C_d$ & 0.560 & $A$ & 10.000 $m^2$ & $h$ & 1.500 m  \\
            $w$ & 2.030 m & $N$ & 5 steps & $\lambda$ & 5 \\
			\bottomrule
		\end{tabular}
    }
	\end{threeparttable}
	\label{table:config}
\end{table}

\subsection{Implementation Details}

\begin{figure}[!t]
    \centering
    \includegraphics[width=\linewidth]{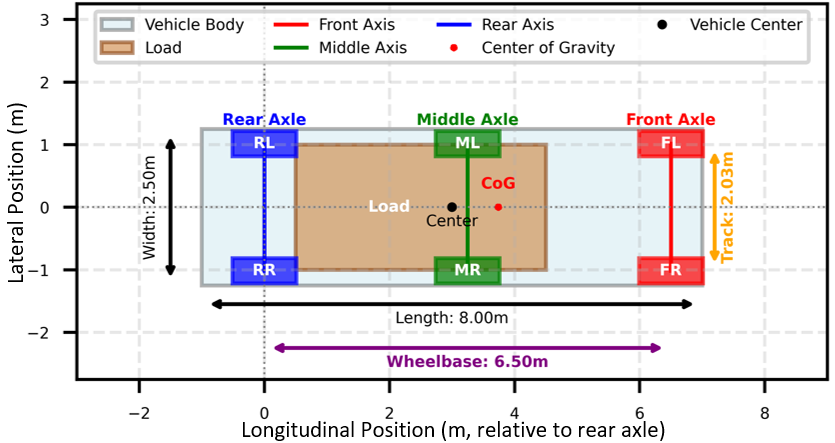}
    \caption{The spatial structure of the implemented DET.}
    \label{fig:blueprint}
\end{figure}

The nuTruck benchmark, along with the reimplemented planners and controllers, are implemented using the PyTorch library. To faithfully reflect the dynamical characteristics of DETs, we align the vehicle dynamical parameters with those of a reference truck, a 3A conventional van with six wheels, three axles, and a 5000 kilograms payload in TruckSim 2019.0 software.
Key configurations of the simulator are summarized in Tab. \ref{table:config}. 
The implemented DET has a length of 8.0 meters and a width of 2.5 meters, whose dimensions are significantly larger than those of typical passenger cars, thereby posing challenges for AD planners. Its spatial structure is illustrated in Fig. \ref{fig:blueprint}.
To balance tracking accuracy and computational efficiency in the simulation, the planners and controllers operate at 10 Hz. The resulting control signals are then interpolated to a high frequency (100 Hz) for vehicle actuation, which also enhances the state propagation accuracy of the dynamical model. 
Each simulation lasts a maximum of 15 seconds following the nuPlan benchmark \cite{nuplan}.

\subsection{Fidelity Evaluation of Ego Dynamics}

\begin{table}[!t]
	\centering
	\caption{The 15-second prediction accuracy of the implemented dynamical model in the nuTruck simulation.}
    \renewcommand\arraystretch{1.5}
	\begin{threeparttable}
    \resizebox{\linewidth}{!}{
		\begin{tabular}{l | c c c c c c | r}
			\toprule
			{State} & $p_x$ & $p_y$ & $\psi_z$ & $v_x$ & $v_y$ & $\omega_z$ & LTR \\
			\midrule
            MAE & 0.25 m & 0.26 m & 0.83 $\degree$ & 0.04 m/s & 0.02m/s & 0.16 $\degree/s$ & 0.005 \\
            RMSE & 0.41 m & 0.37 m & 1.16 $\degree$ & 0.05 m/s & 0.06 m/s & 0.22 $\degree/s$ & 0.017 \\
			\bottomrule
		\end{tabular}
    }
	\end{threeparttable}
	\label{table:dynamics}
\end{table}

\begin{figure}[!t]
    \centering
    \includegraphics[width=\linewidth]{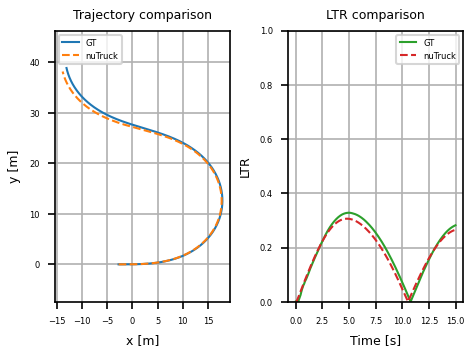}
    \caption{A 15-second case study comparing the dynamical state prediction accuracy between nuTruck and TruckSim simulation.}
    \label{fig:dynamics}
\end{figure}

In the nuTruck simulation, a nonlinear dynamical model is implemented for DETs, which is essential for enabling closed-loop evaluation of dynamics-aware planning. To verify that the implemented model faithfully captures the dynamical characteristics of DETs, we compare it against the ground-truth truck model from TruckSim software. Under identical control inputs and initial dynamical states, we evaluate the prediction accuracy of dynamical states over time using mean absolute error (MAE) and root mean square error (RMSE). A large volume of control signals, simulated as sine curves, is used to collect 124 clips for evaluation. As shown in Tab. \ref{table:dynamics}, the implemented dynamical model predicts state evolution with high accuracy over 15‑second simulations, achieving performance comparable to that of TruckSim. Notably, the model provides accurate estimates of the LTR value, indicating its ability to reliably indicate rollover risk, which serves as a critical feature for the nuTruck benchmark. 

As shown in Fig. \ref{fig:dynamics}, during a complex S‑shaped trajectory, the model maintains high fidelity in describing vehicle motion of DETs and effectively estimates temporal changes in rollover risk caused by dynamic vertical load transfer.

\begin{table*}[!t]
	\centering
	\caption{Closed-loop Performance of existing planners on the nuTruck benchmark.}
    \renewcommand\arraystretch{1.5}
	\begin{threeparttable}
    \resizebox{\linewidth}{!}{
		\begin{tabular}{ l l | c c c c | c c c c | c c c c}
			\toprule
			\multirow{2}{*}{Planner} & \multirow{2}{*}{Controller} & \multicolumn{4}{c|}{Static scenario challenge} & \multicolumn{4}{c|}{Nonreactive agent challenge} & \multicolumn{4}{c}{Reactive agent challenge} \\
            & & CLS & CLS-Safe & NRS & EPAR & CLS & CLS-Safe & NRS & EPAR & CLS & CLS-Safe & NRS & EPAR \\
			\midrule
            \multicolumn{14}{l}{\textbf{Rule-based trajectory planner+controller}} \\
            \hline
            IDM \cite{idm} & iLQR & 75.65 & 49.44 & 62.03 & 85.63 & 59.78 & 39.64 & 63.45 & 77.02 & 63.28 & 41.75 & 63.43 & 76.65 \\
            IDM \cite{idm} & iLQR-S & 73.67 & 59.22 & 74.85 & 83.28 & 57.94 & 47.12 & 75.06 & 74.08 & 62.04 & 50.35 & 75.53 & 73.73 \\
            PDM \cite{pdm} & iLQR & 69.96 & 14.90 & 20.79 & 68.77 & 56.72 & 18.94 & 31.54 & 60.11 & 59.72 & 17.02 & 28.54 & 59.53 \\
            PDM \cite{pdm} & iLQR-S & 70.60 & 23.62 & 32.92 & 74.69 & 59.40 & 26.92 & 44.14 & 65.60 & 61.36 & 27.28 & 42.33 & 64.05 \\
            \hline
            \multicolumn{14}{l}{\textbf{Learning-based trajectory planner+controller}} \\
            \hline
            Plan-R1 \cite{planr1} & iLQR & 77.93 & 8.76 & 9.84 & 82.35 & 73.67 & 8.16 & 9.17 & 88.76 & 71.84 & 7.25 & 9.18 & 84.59 \\
            Plan-R1 \cite{planr1} & iLQR-S & 83.44 & 52.87 & 61.21 & 84.28 & 71.92 & 42.82 & 60.04 & 87.97 & 69.71 & 44.78 & 62.39 & 84.47 \\
            \hline
            \multicolumn{14}{l}{\textbf{Learning-based action planner}} \\
            \hline
            \multicolumn{2}{l|}{CaRL-A12 \cite{carl}} & 75.09 & 67.11 & 89.83 & 68.99 & 68.33 & 55.83 & 81.77 & 70.99 & 70.14 & 59.16 & 85.28 & 69.72 \\
            \multicolumn{2}{l|}{CaRL-A2-FRO \cite{carl}} & 78.80 & 77.36 & 97.67 & 73.71 & 71.09 & 64.44 & 90.38 & 74.62 & 73.44 & 68.17 & 92.85 & 73.25 \\
            \multicolumn{2}{l|}{CaRL-A2-RWL \cite{carl}} & 92.61 & 68.70 & 74.24 & 90.39 &82.47 & 61.99 & 74.84 & 86.35 & 81.73 & 61.58 & 75.17 & 84.87 \\
            \multicolumn{2}{l|}{CaRL-A4 \cite{carl}} & 78.29 & 66.45 & 84.51 & 75.05 & 67.43 & 53.82 & 74.36 & 76.96 & 72.85 & 59.27 & 80.18 & 73.51 \\
			\bottomrule
		\end{tabular}
    }
    \begin{tablenotes}
        \item EPAR score is a metric that measures the progress of the ego vehicle along the route, serving as a key indicator of task completion performance in the nuPlan benchmark \cite{nuplan}.
    \end{tablenotes}
	\end{threeparttable}
	\label{table:main}
\end{table*}

\subsection{Planning Performance Evaluation of Baselines}

To investigate the performance of well-known planners when adapted to DETs, we evaluate the closed-loop planning performance of various baselines, including both rule-based ones \cite{idm,pdm} and 
learning-based ones \cite{planr1,carl}, in both static scenarios (no agents) and dynamic scenarios (nonreactive or reactive agents). The results are listed in Tab. \ref{table:main}. 

Experimental results reveal that although existing well-performing planners can predict collision-free trajectories based on driving scenario information, they fail to account for dynamical safety. This issue is particularly critical for DET planning tasks, since tracking such trajectories imposes a high rollover risk on DETs. The problem is even more pronounced for trajectory-based planners \cite{idm,pdm,planr1}.
In theory, PDM-Closed \cite{pdm} extends IDM \cite{idm} by introducing lateral offsets to sample a wider range of trajectories. For computational efficiency, however, we retain the original kinematic model in the scoring stage rather than employing a full dynamical model. Consequently, the lateral offset induces larger lateral accelerations, which aggravate dynamic load transfer and significantly reduce the NRS score.
Plan-R1 \cite{planr1} incorporates closed-loop RL-based fine-tuning after IL-based pre-training, delivering enhanced EPAR scores, yet its NRS score remains low due to the absence of dynamical characteristics during its fine-tuning stage.

Even when rollover prevention is considered in the trajectory tracking controller, the trajectories cannot be fully optimized since the planner neglects dynamical safety, making it difficult to completely mitigate rollover risk. Moreover, the iLQR-S controller, in its effort to avoid rollover, often fails to accurately track the collision-free trajectories provided by the planner. This compromises passability in certain complex scenarios and consequently leads to occasional degradation of the EPAR score.

For action planners \cite{carl}, the neural network directly predicts low‑level actions. Although we do not explicitly incorporate rollover prevention in the RL-based training, the smoothness of the neural network, together with the driving comfort encouraged by the reward function, yields a significantly higher NRS score than that of trajectory planners \cite{idm,pdm,planr1}. In addition, CaRL \cite{carl} does not require expert trajectory data, which facilitates its training for DET configurations. Under the combined evaluation framework that accounts for both collision‑free and rollover‑free performance, CaRL \cite{carl} achieves the best overall results.

These results support our claim. For DET planning tasks in complex scenarios, the conventional paradigm, in which the planner ensures collision avoidance and the controller handles rollover prevention, falls short of achieving an ideal balance between kinematic passability and dynamical safety.

\subsection{Further Analysis about Action Planner}

We further investigate the influence of action patterns and training steps on action planner performance, with the goal of assessing whether the scalability established on passenger cars \cite{carl} generalizes to DETs.

\begin{figure}[!t]
    \centering
    \includegraphics[width=\linewidth]{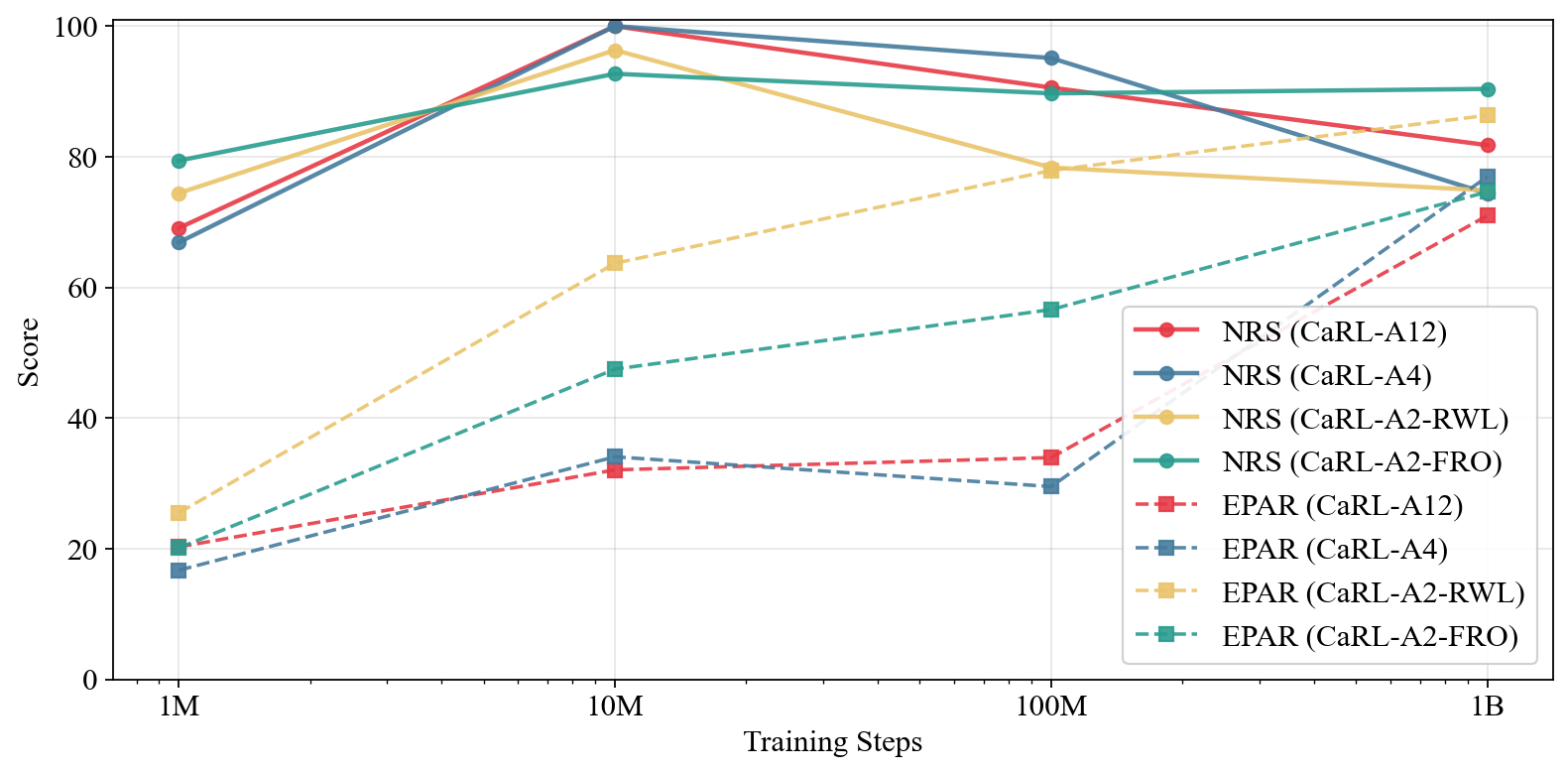}
    \caption{The comparison between CaRL planners \cite{carl} trained under different patterns and training steps.}
    \label{fig:scaling}
\end{figure}

We first compare vanilla CaRL-A12 \cite{carl} with its variants with action patterns. As shown in Tab. \ref{table:main}, after the same number of training steps, all CaRL variants \cite{carl} with action patterns achieve improved EPAR scores, demonstrating the effectiveness of action patterns.
In addition, as illustrated in Fig. \ref{fig:scaling}, the convergence speeds of CaRL-A12 \cite{carl} and CaRL‑A4 \cite{carl} are notably slower than that of CaRL‑A2 \cite{carl}. This is because the high complexity of the DET action space makes it difficult for the network to directly learn high‑dimensional control signals, leading to slow convergence. During such prolonged training, the network struggles to generate reasonable control signals, such as reverse actuation between wheels at the front and rear axles. The FRO and RWL action patterns effectively constrain the predicted control signals within a reasonable pattern, thereby facilitating network convergence.

Compared with CaRL-A2-FRO \cite{carl}, the variant with RWL pattern achieves higher route completion. However, to traverse the same scenario, CaRL‑A2‑RWL \cite{carl} generally requires larger steering angles than CaRL‑A2‑FRO \cite{carl}, which results in more aggressive lateral acceleration changes and consequently a higher rollover risk.

Fig. \ref{fig:scaling} also confirms the scalability of the RL‑based action planner for DET planning, as route completion steadily improves with training steps. However, rollover risk increases in late training stages due to high‑speed traversals, highlighting the need for considering dynamical safety in RL training.

\section{Conclusions}

In this letter, we explore an evaluation benchmark for AD planners tailored to DET. Given that existing datasets and benchmarks do not incorporate dynamical safety as an evaluation metric, and that DETs have a relatively high rollover risk, we propose the nuTruck benchmark. By integrating real-world driving scenario data from nuPlan dataset \cite{nuplan} with a high-fidelity DET dynamical model, our benchmark enables real-time monitoring of dynamical properties during both training and evaluation of AD planners, thereby allowing quantitative assessment of dynamical safety. On this newly constructed benchmark, we evaluate several state-of-the-art planning approaches, including rule-based and learning-based methods. Experimental analyses demonstrate that conventional trajectory planners commonly used for passenger cars are not directly applicable to DETs, and highlight that DET planners should consider rollover prevention in addition to collision avoidance. We further analyze the scalability issues in training learning-based planning models, and find that incorporating action pattern priors of DETs effectively facilitates model convergence. The nuTruck benchmark proposed in this work can serve as a new evaluation standard for future research on DET-oriented AD planning, and the experimental findings presented herein can provide valuable references for subsequent studies.




\end{document}